\documentclass[letterpaper, 10 pt, journal, twoside]{IEEEtran}

\IEEEoverridecommandlockouts 

\usepackage{graphics} 
\usepackage{epsfig} 
\usepackage{times} 
\usepackage{amsmath} 
\usepackage{amssymb}  
\usepackage{hyperref} 
\usepackage{xcolor} 
\usepackage{arydshln} 
\usepackage{color} 
\usepackage{float}
\usepackage{multirow}
\usepackage{subcaption}
\usepackage{bm}

\newcommand{\rebuttal}[1]{{\color{blue} #1}}
\newcommand\norm[1]{\left\lVert#1\right\rVert}
\newcommand{\reffig}[1]{Fig.~\ref{fig:#1}}

\newcommand{\qmarks}[1]{``#1''}

\title{
Continuous-Time vs. Discrete-Time Vision-based SLAM: A Comparative Study
}

\author{Giovanni Cioffi, Titus Cieslewski, and Davide Scaramuzza
\thanks{Manuscript received: September, 9, 2021; Revised December, 1, 2021; Accepted December, 27, 2021.}

\thanks{This paper was recommended for publication by Editor Javier Civera upon evaluation of the Associate Editor and Reviewers' comments.

This work was supported by the National Centre of Competence in Research (NCCR) Robotics through the Swiss National Science Foundation (SNSF) and the European Union’s Horizon 2020 Research and Innovation Programme under grant agreement No. 871479 (AERIAL-CORE) and the European Research Council (ERC) under grant agreement No. 864042 (AGILEFLIGHT).}

\thanks{The authors are with the Robotics and Perception Group, Department of Informatics, University of Zurich, and Department of Neuroinformatics, University of Zurich and ETH Zurich, Switzerland (\protect\url{http://rpg.ifi.uzh.ch}). {\tt\footnotesize cioffi@ifi.uzh.ch}}%
\thanks{Digital Object Identifier (DOI): see top of this page.}
}

\begin{document}

\maketitle


\begin{abstract}
Robotic practitioners generally approach the vision-based SLAM problem through discrete-time formulations.
This has the advantage of a consolidated theory and very good understanding of success and failure cases.
However, discrete-time SLAM needs tailored algorithms and simplifying assumptions when high-rate and/or asynchronous measurements, coming from different sensors, are present in the estimation process. 
Conversely, continuous-time SLAM, often overlooked by practitioners, does not suffer from these limitations. 
Indeed, it allows integrating new sensor data asynchronously without adding a new optimization variable for each new measurement. 
In this way, the integration of asynchronous or continuous high-rate streams of sensor data does not require tailored and highly-engineered algorithms, enabling the fusion of multiple sensor modalities in an intuitive fashion.
On the down side, continuous time introduces a prior that could worsen the trajectory estimates in some unfavorable situations.
In this work, we aim at systematically comparing the advantages and limitations of the two formulations in vision-based SLAM.
To do so, we perform an extensive experimental analysis, varying robot type, speed of motion, and sensor modalities.
Our experimental analysis suggests that, independently of the trajectory type, continuous-time SLAM is superior to its discrete counterpart whenever the sensors are not time-synchronized.
In the context of this work, we developed, and open source, a modular and efficient software architecture containing state-of-the-art algorithms to solve the SLAM problem in discrete and continuous time.
\end{abstract}


\section*{Supplementary Material} \label{sec:SupplementaryMaterial}

The code can be found here~\url{https://github.com/uzh-rpg/rpg_vision-based_slam}
\section{Introduction} \label{sec:Introduction}

\IEEEPARstart{S}{imultaneous} localization and mapping (SLAM)~\cite{Cadena16tro} is the problem of building a map of the environment and concurrently estimating the state of the robot.
Accurate and robust SLAM algorithms are the keys to unlock autonomous robotic navigation.
\begin{figure}[H]
    \centering
    \begin{subfigure}[t]{\columnwidth}
        \includegraphics[trim=0 250 0 0, clip, width=\columnwidth]{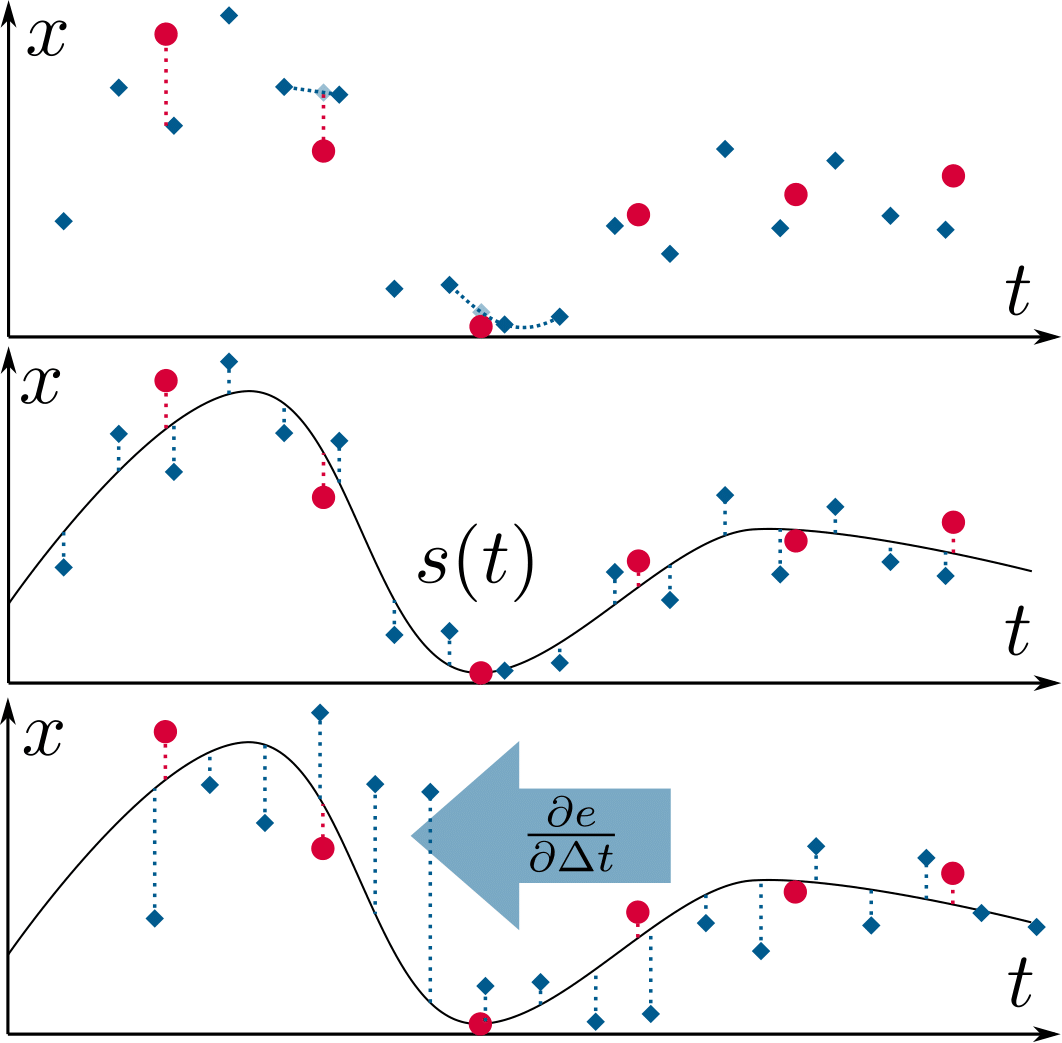}
        \caption{It is typical among popular SLAM methods to represent the state discretely at the measurement times of one of the sensors, e.g., the camera in vision-based SLAM.
        If the other sensor does not have measurements at the same times, techniques such as interpolation need to be employed to express error terms.}
    \end{subfigure}
    \begin{subfigure}[t]{\columnwidth}
        \includegraphics[trim=0 125 0 125, clip, width=\columnwidth]{images/eyecatcher.png}
        \caption{In continuous-time SLAM, the estimated state is instead expressed using a continuous function, e.g., a spline $s(t)$.
        Now, for any measurement at a time $t_i$, a meaningful error term can be expressed by comparing the measurement to the spline sample $s(t_i)$, or any of its derivatives $s'(t_i), s''(t_i), ...$, e.g., no integration needed for IMU measurements.}
    \end{subfigure}
    \begin{subfigure}[t]{\columnwidth}
        \includegraphics[trim=0 0 0 250, clip, width=\columnwidth]{images/eyecatcher.png}
        \caption{Furthermore, a continuous-time representation allows the simultaneous estimation of time offset or even drift between sensors.
        To this end, the error terms of one of the sensors is simply expressed with the spline sampled at its measurement times plus a constant time offset shared among all measurements, $s(t_i + \Delta t)$.
        Then, $\Delta t$ can be co-optimized with the parameters of $s$.
        In this example, the blue measurements would thus be \qmarks{shifted back to the left}.}
    \end{subfigure}
    \caption{Benefits of using a continuous-time state representation illustrated on a simple example where a variable $x(t)$ is estimated from two noisy sensors that measure $x$ at different frequencies (red dots and blue diamonds).}
    \label{fig:eyecat}
\end{figure}

Among the plethora of sensors providing relevant information for localization and mapping, cameras are a very convenient solution in virtue of their information-rich measurements, low cost, and low weight.
Vision-based SLAM algorithms use cameras as the main, but not necessarily the only sensor modality.
The most common vision-based SLAM formulation is based on the discrete-time (DT) trajectory representation~\cite{durrant2006simultaneous}.
Namely, the poses of the camera are estimated at the time a new measurement, i.e., an image, is available.
The discrete-time formulation has the benefit of a very consolidated theory.
In the past years, many successful applications have been seen~\cite{Cadena16tro}.
Its limitations are well understood and are addressed in ongoing research. 

Although cameras can be used as the only source of information in SLAM systems, fusing multiple sensor modalities is beneficial for accuracy and robustness.
In discrete-time SLAM, customized algorithms are necessary to include asynchronous measurements coming from different sources in the estimation process~\cite{qin2018online}.
Similarly, ad-hoc solutions are needed to avoid adding a new state to the estimation problem every time a new measurement is available~\cite{Forster17troOnmanifold}.
In this way, high-rate sensors can be incorporated, thus limiting the increase in computational complexity.

In the past years, researchers have been investigating the use of continuous-time (CT) representations to encode the camera trajectory~\cite{Furgale12icra,ovren2019trajectory, Mueggler18tro}.
Continuous-time based probabilistic SLAM formulations have been derived similarly to the discrete-time case.
In particular, applications of continuous-time trajectory representations have been used for multi-sensors calibration~\cite{Furgale12icra,Furgale13iros}, planning~\cite{ding2019efficient}, and 3D reconstruction~\cite{yang2021asynchronous,ovren2018spline}.
The continuous-time formulation brings several advantages to the estimation problem.
Firstly, continuous-time trajectories can be sampled at any time.
This makes it easy to fuse asynchronous sensors and estimate time offsets, see~\reffig{eyecat}.
Such property is also beneficial for sensors with continuous stream of data, e.g., LiDARs, rolling shutter cameras, and event cameras.
Secondly, the continuous-time formulation removes the need to include an optimization variable for every sensor measurement.
The computational complexity of the optimization problem is kept bounded, allowing to easily include high-rate sensors, such as inertial measurement units (IMU), in the estimation process.
However, the continuous-time representation introduces a prior on the smoothness of the trajectory.
Modeling this prior such that it can generalize to different levels of the trajectory smoothness is not an easy task.
For example, when using polynomial functions for the continuous-time representation, a not high enough degree of the polynomial function would lead to a loss of details in the trajectory estimates due to excessive smoothing.

To the best of our knowledge, there is no systematic comparison between the continuous- and discrete-time formulations for vision-based SLAM.
Such systematic analysis is fundamental to guide the robotic practitioners in the design of future SLAM solutions.
Therefore, we perform an extensive quantitative analysis to understand the respective advantages and limitations of the two trajectory representations.
We focus on batch SLAM with visual, inertial, and global positional (i.e., Global Positioning System (GPS)) measurements and also investigate the contribution of each sensor modality.
IMU and GPS provide local high-rate and global low-rate measurements, respectively, which are complementary to the camera measurements.
Camera, IMU, and GPS measurements are fused together to obtain locally accurate and long-term drift-free trajectory estimates.
We run experiments in both hardware-in-the-loop simulation and on real-world trajectories of flying and ground robots.

Our experiments indicate that discrete-time and continuous-time representations produce equivalent results when the sensors are time-synchronized.
However, when there is an offset in the time synchronization, continuous-time is superior.
The main reason behind this result is that the simplifying assumptions made for estimating the time offsets in discrete time do not always hold.
These finding are valid for both aerial and ground robots.

To summarize, the main contributions of this work are:
\begin{itemize}
    \item An in-depth study on the comparison of discrete- and continuous-time trajectory representations in vision-based SLAM.
    \item Extensive experimental evaluation in hardware-in-the-loop simulation and on real-world data collected from flying and ground robots.
    \item A modular, efficient software architecture including state-of-the-art algorithms to solve the SLAM problem in the discrete and continuous time. We release the code fully open-source.
\end{itemize}
\section{Related Work} \label{sec:RelatedWork}

A popular choice for approximating continuous functions are temporal basis functions.
In the set of temporal basis functions, B-splines~\cite{Furgale12icra, sommer2020efficient} have properties that are useful for trajectory representation: \textit{locality} and \textit{smoothness}.
B-splines define the continuous time evolution by using a set of control points.
Sampling at any time only depends on a local subset of the control points (\textit{locality}).
The size of this subset corresponds to the order $k$ of the B-spline.
A B-spline of order $k$ is a function of class $C^{k-1}$ (\textit{smoothness}).
Other suitable choices of basis functions are wavelets~\cite{Anderson14icra}.
Gaussian processes have also been used to represent continuous-time trajectories for batch state estimation problems~\cite{anderson2015batch}.\\
Recent works have tried to reduce the computational complexity of sampling and computing derivatives from B-splines.
In~\cite{sommer2020efficient}, a novel derivation for the computation of the time derivatives of cumulative B-spline is proposed.
This approach reduces the computational complexity from quadratic to linear in the order of the spline.
A novel efficient non-uniform split interpolation for cumulative B-splines is proposed in~\cite{hug2020conceptualizing}.\\
The most related applications of continuous-time trajectory representation to our work are multi-sensor calibration~\cite{Furgale12icra, Furgale13iros}, 3D reconstruction~\cite{ovren2019trajectory, yang2021asynchronous}, visual-inertial odometry for event cameras~\cite{Mueggler18tro}, and batch state estimation~\cite{anderson2015batch}.

The literature on the classical discrete-time SLAM formulation is very broad.
Due to space constraints, we refer the reader to the survey paper~\cite{Cadena16tro} for an overview on the discrete SLAM formulation.
Differently from continuous time, specific algorithms are needed to estimate the time offsets between different sensors.
In our discrete-time implementation, we use the method proposed in~\cite{qin2018online} to estimate the camera-IMU time offset.
A similar approach to the one in~\cite{lee2020intermittent} is used for the GPS-IMU time offset estimation.
We describe these two methods in more details in Sec.~\ref{sec:Methodology_Discrete_time}.

\begin{figure}[H]
    \centering
    \begin{tabular}{cc}
        \includegraphics[width=0.45\linewidth]{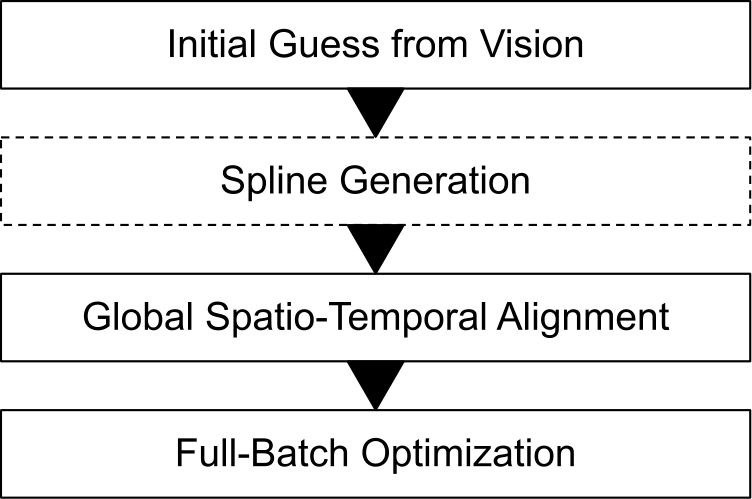}
        &\includegraphics[width=0.45\linewidth]{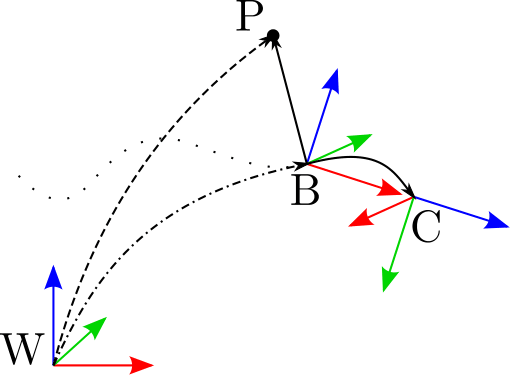}\\
        (a) Estimation flowchart & (b) Reference frames
    \end{tabular}
    \caption{(a) The steps involved in the batch trajectory estimation.
    Spline generation only applies in continuous-time estimation.\\
    (b) The frames and the positions involved in the estimation process.
    Solid \rebuttal{black} lines indicate optimizable, calibrated transformations, the dashed lines the position measurements, and the dashed-dotted lines the time-varying, estimated transformation.
    The dotted lines indicate the trajectory of the latter.
    {\color{red}Red}, {\color{green}green} and {\color{blue}blue} arrows indicate {\color{red}x}-, {\color{green}y}- and {\color{blue}z}-axes of the frames.
    }
    \vspace{-2ex}
    \label{fig:flowchart_and_ref_frames}
\end{figure}

\section{Methodology} \label{sec:Methodology}

We solve the estimation problem using a multi-step approach that involves a few initialization steps before the full-batch optimization.
Fig.~\ref{fig:flowchart_and_ref_frames} (a) shows the flow chart of the optimization pipeline for the both continuous- and discrete-time approaches.
The reference frames are depicted in Fig.~\ref{fig:flowchart_and_ref_frames} (b).
$W$ is the fixed world frame, whose $z$ axis is aligned with the gravity. 
When available, GPS measurements are expressed in this frame. $B$ is the moving body frame. 
We set it equal to the IMU frame. 
$C$ is the camera frame. 
$P$ is the GPS antenna position. Note that this is not a full 6-DOF frame, as no orientation is provided by the GPS measurement.
We use the notation $(\cdot)^{w}$ to represent a quantity in the world frame $W$. Similar notation applies for each reference frame.
The position, orientation, and velocity of $B$ with respect to $W$ at time $t_k$ are written as $\mathbf{p}_{b_{k}}^{w} \in \mathbb{R}^3$, $\mathbf{R}_{b_{k}}^{w} \in \mathbb{R}^{3 \times 3}$ part of the 3-D rotation group $SO(3)$, and $\mathbf{v}_{b_{k}}^{w} \in \mathbb{R}^3$, respectively.
We use 4$\times$4 matrices, $\mathbf{T} \in SE(3)$ (the special Euclidean group) to express 6-DOF Euclidean transformations.\\
GPS measurements consist in the position of the GPS antenna in the world frame, $\mathbf{p}_{p}^{w} \in \mathbb{R}^3$.
The position of the $i$-th 3-D visual landmark in the world frame is $\mathbf{p}_{l_{i}}^{w} \in \mathbb{R}^3$.
The set $\mathcal{L}$ contains the 3-D visual landmarks.
The time $t^{c}_{i}$ is the time offset between camera and IMU such that $t_{imu} = t_{cam} + t^{c}_{i}$. 
Using the same convention, $t^{g}_{i}$ is the GPS-IMU time offset.\\
The state vectors containing the optimization variables are $^{c}\mathcal{X}$ and $^{d}\mathcal{X}$, for the continuous- and discrete-time approaches, respectively.
They are introduced in Sec.~\ref{sec:Methodology_Continuous_time} and Sec.~\ref{sec:Methodology_Discrete_time}.
The transformation $\mathbf{T}^{c}_{i}$ is the extrinsic calibration matrix between camera and IMU.
The vector $\mathbf{p}_{p}^{b} \in \mathbb{R}^3$ is the position of the GPS antenna in the body frame.
We use different representations for the IMU biases in continuous time, $^{c}\mathcal{B}$, and discrete time, $^{d}\mathcal{B}$. They are introduced in Sec.~\ref{sec:Methodology_Continuous_time} and Sec.~\ref{sec:Methodology_Discrete_time}, respectively.

In this work, we focus on global shutter cameras, which are the common camera choice in vision-based SLAM systems.
The initial camera poses and 3-D landmarks are obtained by using COLMAP~\cite{Schoenberger16cvpr}.
COLMAP is a monocular Structure-from-Motion pipeline which is widely used in the computer vision and robotic community.
The camera poses and 3-D landmarks reconstructed by COLMAP are expressed in the scaleless reference frame $G$, which is a fixed frame defined such that the first camera pose is equal to the identity.
We use the notation $\bar{\cdot}$ to denote a sensor measurement and the notation $\hat{\cdot}$ to denote a ground-truth measurement.

\subsection{Continuous-time representation}\label{sec:Methodology_Continuous_time}

We use cumulative B-splines for the continuous-time trajectory representation.
A uniform B-spline of order $k$ and $N+1$ control nodes $\mathbf{x}_{i}$ is defined by 
\begin{equation}\label{eq:bspline_eq}
    \mathbf{x}(t) = \sum_{i=0}^{N} B_{i,k}(t) \mathbf{x}_{i},
\end{equation}
where $t$ is the time variable.
The control points are equally spaced in time by an interval $\Delta t$, $t_i = t_0 + i \Delta t$, and $B_{i,k}(t)$ are given by the De Boor-Cox recurrence relations~\cite{DeBoor01splines}.
For cumulative B-splines, Eq.~(\ref{eq:bspline_eq}) can be rewritten as
\begin{equation}
    \mathbf{x}(t) = \tilde{B}_{0,k}(t) \mathbf{x}_0 + \sum_{i=0}^{N} \tilde{B}_{i,k}(t) \mathbf{d}_{i},
\end{equation}
with $\tilde{B}_{i,k}(t) = \sum_{s=i}^N B_{s,k}(t)$, and $\mathbf{d}_i = \mathbf{x}_i - \mathbf{x}_{i-1}$.
The coefficients of a uniform B-spline are constant and can be written in matrix form~\cite{Qin00vc}. 
This matrix form can be also used for cumulative B-splines.
Combining the matrix formulation~\cite{Qin00vc} with the uniform representation proposed in~\cite{sommer2020efficient}, the calculations can be simplified and the equation for sampling from the B-spline at time $t$ becomes
\begin{equation}\label{eq:cumulative_bspline}
    \mathbf{x}(u) = \mathbf{x}_i + \sum_{j=1}^{k-1} \lambda_j (u) \cdot \mathbf{d}_j^i.
\end{equation}
The uniform time representation proposed in~\cite{sommer2020efficient} defines $u(t)$, with $t \in [t_i, t_{i+1})$, as the normalized time elapsed since the start of the segment. 
Defined $h(t) = \frac{t-t_0}{\Delta t}$, $u$ can be computed as $u(t) = h(t) - i$.
The coefficients $\lambda_j (u) \in \mathbb{R}$ are constant and only depend on the order of the B-spline (see Eq. (18-21) in~\cite{sommer2020efficient}).
The difference vector is $\mathbf{d}_j^i = \mathbf{x}_{i+j} - \mathbf{x}_{i+j-1}$.
In~\cite{sommer2016continuous}, the cumulative B-spline formulation was derived for elements of Lie groups.
In the Lie group $SO(3)$, the addition corresponds to the matrix multiplication, but scaling by a scalar $\lambda$ is not defined.
The scaling operation requires to map an element, $\mathbf{R} \in SO(3)$, of the group to a vector space (the Lie algebra), performing the scaling operation, and then remapping to the Lie group: $\text{Exp}(\lambda \cdot \text{Log}(\mathbf{R}))$.
Elements of the Lie algebra can be mapped to the Lie group by the exponential map, Exp($\cdot$). 
The inverse operation is the logarithm map, Log($\cdot$).
The cumulative B-spline formulation for $SO(3)$ is
\begin{equation}
    \mathbf{R}(u) = \mathbf{R}_i \cdot \prod_{j=1}^{k-1} \text{Exp}(\lambda_j (u) \cdot \mathbf{d}_j^i),
\end{equation}
where $\mathbf{d}_j^i = \text{Log}(\mathbf{R}_{i+j-1}^{-1} \cdot \mathbf{R}_{i+j})$.
We use two B-splines to represent the position, $\mathbf{p}(u) \in \mathbb{R}^3$, and orientation, $\mathbf{R}(u) \in SO(3)$, of the trajectory we are interested in estimating.
We use the notation $^{c}\mathcal{T} = \{\mathbf{p}_i, \mathbf{R}_i \}$, with $i = 0, \cdots , N$, and $N+1$ the number of control nodes, to denote the continuous-time representation of a 6-DOF trajectory.
Using one ($^{c}\mathcal{T} = \{ \mathbf{T}_i \}, \mathbf{T} \in SE(3)$) or two splines to represent a 6-DOF trajectory is equivalent, but the two splines solution is computationally less expensive~\cite{sommer2020efficient,haarbach2018survey, ovren2019trajectory}.
We compute the time derivatives of the B-spline as proposed in~\cite{sommer2020efficient}, which reduces the number of matrix operations from $\mathcal{O}(k^2)$ to $\mathcal{O}(k)$.

\subsubsection{Initialization}\label{sec:Methodology_continuous_time_initialization}

The first step of our continuous-time trajectory estimation pipeline is to fit a B-spline to the $K$ camera poses estimated by COLMAP: $\mathbf{p}^{g}_{c_k}, \mathbf{R}^{g}_{c_k}$.
The initial values of the $N$ control nodes, $\mathbf{p}^{g}_{c_i}$ and $\mathbf{R}^{g}_{c_i}$, are obtained by linearly interpolating the COLMAP camera poses at the times $t_i = t_0 + i \Delta t$, where $t_0$ is the time of the first camera pose and $\Delta t$ is the inverse of the control node frequency.
Then, the control nodes are optimized to minimize the cost function
\begin{equation}\label{eq:cost_fun_spline_fit_to_colmap}
    \min_{\mathbf{p}^{g}_{c_i}, \mathbf{R}^{g}_{c_i}} \sum_{j = 1}^M \norm{\mathbf{p}^{g}_{c}(u(t_j)) - \mathbf{p}^{g}_{c_j}}^2 + \norm{\text{Log}(\mathbf{R}^{g}_{c}(u(t_j))^t \cdot \mathbf{R}^{g}_{c_j})}^2,
\end{equation}
where $M$ is the number of errors. 
The measurements $\mathbf{p}^{g}_{c_j}$ are obtained by the linear interpolation of the camera poses $\mathbf{p}^g_{c_k}$ and $\mathbf{p}^g_{c_{k+1}}$, with $t_j \in [t_k, t_{k+1})$.
We similarly use spherical linear interpolation (SLERP) for the rotations $\mathbf{R}^{g}_{c_j}$.
We minimize the cost function till convergence, which usually happens in less than 20 iterations. 
The result of this step is the continuous-time trajectory $^{c}\mathcal{T}^g_c = \{ \mathbf{p}^{g}_{c_i}$, $\mathbf{R}^g_{c_i} \}$ that represents the camera poses in the reference frame $G$.
We transform the camera poses in body poses and obtain the trajectory $^{c}\mathcal{T}^g_b = \{ \mathbf{p}^g_{b_i}$, $\mathbf{R}^g_{b_i} \}$.
The initial value of $\mathbf{T}^{c}_{i}$ is obtained by using the calibration toolbox Kalibr~\cite{Furgale13iros}.

The second step of our continuous-time trajectory estimation pipeline is to estimate the actual scale of the trajectory $^{c}\mathcal{T}^g_b$ as well as to find a transformation that aligns it to the gravity aligned frame.
When GPS measurements are available, we obtain an initial estimation of the 6-DOF transformation $\mathbf{T}^w_g$ and scale $s$ using the method proposed in~\cite{Umeyama91pami}.
This method finds the least-squares solution that minimizes, with respect to $\mathbf{T}^w_g$ and $s$, the differences between $\mathbf{p}^g_b(u(t_j))$, sampled from the spline, and the GPS measurements $\bar{\mathbf{p}}^w_{p_j}$.
Here we assumes that the GPS antenna-IMU positional offset, $\mathbf{p}^b_p$, is null.
Then, $\mathbf{T}^w_g$, $s$, $\mathbf{p}^b_p$, and $t^{g}_{i}$, are estimated by minimizing the cost
\begin{equation}\label{eq:cost_fun_align_spline_to_global_frame}
    \min_{\mathbf{T}^w_g, s, \mathbf{p}^b_p, t^{g}_{i}} \sum_{j = 1}^D \norm{\bar{\mathbf{p}}^{w}_{p_j} - \mathbf{p}^{w}_{p_j} }^2,
\end{equation}
where the $j$-th $\in [1, \cdots, D)$ GPS measurement, $\bar{\mathbf{p}}^{w}_{p_j}$, is available at time $t_j$. 
The predicted antenna position is sampled from the spline as: $\mathbf{p}^{w}_{p_j} = s \mathbf{R}^{w}_{b_j} \mathbf{p}^b_p + \mathbf{p}^{w}_{b_j}$, $\mathbf{R}^{w}_{b_j} = \mathbf{R}^{w}_{g} \mathbf{R}^{g}_{b}(u(t_j + t^{g}_{i}))$, and $\mathbf{p}^{w}_{b_j} = s \mathbf{R}^{w}_{g} \mathbf{p}^{g}_{b}(u(t_j + t^{g}_{i})) + \mathbf{p}^{w}_{g}$.

In the case when we do not use GPS measurements, we leverage the IMU measurements to obtain an initial estimate of the scale.
We integrate the IMU measurements for a short period of time, usually few seconds, to obtain a small trajectory segment.
This trajectory is expressed in a gravity aligned frame, $I$, which is estimated from the accelerometer measurements collected when the IMU is static.
Similarly as before, we use~\cite{Umeyama91pami} to obtain the transformation $s, \mathbf{T}^i_g$. 
This transformation is applied to transform $^{c}\mathcal{T}^g_b$ to the frame $I$.

\subsubsection{Full-batch optimization}\label{sec:Methodology_continuous_time_full_batch_opt}

We use $\mathbf{T}^w_g$ and $s$ estimated in the initialization step to transform the trajectory $^{c}\mathcal{T}^g_b$ to the global frame $W$ (or $I$ in the case when GPS is not used): $^{c}\mathcal{T}^{w}_{b} = \{ \mathbf{p}^{w}_{b_i}, \mathbf{R}^{w}_{b_i} \}$. 
Similarly, the 3-D landmarks $\mathbf{p}^{g}_{l_r}$ are also transformed to $W$: $\mathbf{p}^{w}_{l_r} = s \mathbf{R}^{w}_{g} \mathbf{p}^{g}_{l_r} + \mathbf{p}^{w}_{g}$.
In the full-batch optimization, the state vector $^{c}\mathcal{X} = \{ ^{c}\mathcal{T}^{w}_{b}, \mathcal{L}, t^{c}_{i}, \mathbf{T}^{c}_{i}, t^{\text{g}}_{i}, \mathbf{p}_{p}^{b}, \mathbf{g}^w, ^{c}\mathcal{B}\}$, is estimated by minimizing the cost function
\begin{align}\label{eq:cost_fun_full_batch_opti_cont_time}
    \min_{^{c}\mathcal{X}} & 
    \sum_{k = 1}^{K} \sum_{r \in \mathcal{R}_k} \norm{\mathbf{e}^{\text{v}}_{k,r}}^{2}_{\mathbf{W}_\text{v}} + 
    \sum_{m = 1}^{M} (\norm{\mathbf{e}^{\text{a}}_{m}}^{2}_{\mathbf{W}_\text{a}} + \norm{\mathbf{e}^{\omega}_{m}}^{2}_{\mathbf{W}_\text{w}}) + \nonumber \\&
    \sum_{d = 1}^{D} \norm{\mathbf{e}^{\text{gps}}_{d}}^{2}_{\mathbf{W}_\text{gps}} + 
    \sum_{f = 1}^{F} 
    (\norm{\mathbf{e}^{\text{b}_{\text{a}}}_{f}}^{2}_{\mathbf{W}_{\text{b}_{\text{a}}}} + 
    \norm{\mathbf{e}^{\text{b}_{\omega}}_{f}}^{2}_{\mathbf{W}_{\text{b}_{\omega}}}).
\end{align}
Optimizing this cost function results in the maximum a posteriori (MAP) estimation of the state vector $^{c}\mathcal{X}$, which is derived by using the probabilistic continuous SLAM formulation and the Gaussian distribution for all the measurements~\cite{Furgale12icra}.
The error $\mathbf{e}^{\text{v}}_{k,r} = \bar{\mathbf{z}}^{k,r} - \pi(\mathbf{R}^c_w (u) \mathbf{p}_{l_r}^{w} + \mathbf{p}^c_w (u))$ is the visual residuals, which describes the re-projection error of the landmark $\mathbf{p}_{l_r}^{w}$.
The camera pose in the world frame is obtained by sampling and inverting the position and orientation from the spline at $t_k + t^{c}_{i}$: $\mathbf{R}^w_c(u) = \mathbf{R}^w_b(u(t_{k} + t^{c}_{i})) \mathbf{R}^{i}_{c}$, $\mathbf{p}^w_c(u) = s \mathbf{R}^w_b(u(t_{k} + t^{c}_{i})) \mathbf{p}^{i}_{c} + \mathbf{p}^w_b(u(t_{k} + t^{c}_{i}))$.
The set $\mathcal{R}_k$ contains all the landmarks that project to the frame $k$.
The image feature measurements $\bar{\mathbf{z}}^{k,r}$ are obtained from COLMAP.
The function $\pi(\cdot)$ denotes the camera projection model and $t_{k}$ is the timestamp of the image.\\
The value $\mathbf{e}^{\text{a}}_{m} = \mathbf{R}^w_b(u(t_{m}))^{-1} (\ddot{\mathbf{p}}^w_b(u(t_{m})) + \mathbf{g}^w) - \bar{\mathbf{a}}_m + \mathbf{b}_a(u(t_{m}))$ is the $m$-th accelerometer residual with respect to the measurement $\bar{\mathbf{a}}_m$.
The quantity $\ddot{\mathbf{p}}^w_b(u(t_{m}))$ is the second derivative of the B-spline encoding the trajectory positions.
The vector $\mathbf{g}^w$ represents the gravity.
We use cubic B-splines to represent accelerometer and gyroscope biases, $\mathbf{b}_a(u)$ and $\mathbf{b}_{\omega}(u)$ as in~\cite{Furgale13iros}.
The $m$-th gyroscope residual is $\mathbf{e}^{\omega}_{m} = \bm{\omega}(u(t_{m})) - \bar{\bm{\omega}}_m + \mathbf{b}_{\omega}(u(t_{m}))$.
We refer the reader to Eq. (38) in~\cite{sommer2020efficient} for the formula to derive $\bm{\omega}(u)$.
The errors $\mathbf{e}^{\text{b}_{\text{a}}}_{f}$ and $\mathbf{e}^{\text{b}_{\omega}}_{f}$ are residuals on the rate of the bias changes.
The GPS errors $\mathbf{e}^{\text{gps}}_{d}$ are the same as the ones defined in Eq.~(\ref{eq:cost_fun_align_spline_to_global_frame}).
The matrices $\mathbf{W}$ are the weights of the residuals.
Their entries are derived from the sensor noise characteristics.
We assume Gaussian noise with standard deviation of 1 pixel for the 2-D image features.

\subsection{Discrete-time representation}\label{sec:Methodology_Discrete_time}

In the discrete-time formulation, the trajectory is represented by the body poses at the rate of the camera: $^{d}\mathcal{T}^w_b = \{ \mathbf{p}^{w}_{b_k}, \mathbf{R}^{w}_{b_k} \}$.
The camera-IMU time offset is estimated using the method proposed in~\cite{qin2018online}, which is integrated in VINS-Mono~\cite{Qin18tro}, a state-of-the-art visual-inertial odometry pipeline.
This method proposes to shift the 2-D image features to account for the time offset between camera and IMU measurements.
It makes the assumption that the camera motion has constant velocity in a short period of time (e.g., between consecutive frames), and, based on this assumption, it calculates the velocity of the 2-D features on the image plane.
This velocity is then used to shift the feature position in the small period of time that corresponds to the camera-IMU time delay (see Eq. (4) in~\cite{qin2018online}).
This allows to include the time offset in the estimation process.
To define the GPS errors, the trajectory is interpolated at the time of the GPS measurements.
The IMU-GPS time offset is taken into account in the interpolation factor similarly as in~\cite{lee2020intermittent}.

\subsubsection{Initialization}\label{sec:Methodology_discrete_time_initialization}

Similarly to Sec.~\ref{sec:Methodology_continuous_time_initialization}, we compute the body poses from the camera poses estimated by COLMAP and then transform them to the world frame $W$ using the 6-DOF and scale transformation obtained by applying~\cite{Umeyama91pami}.

\subsubsection{Full-batch optimization}\label{sec:Methodology_discrete_time_full_batch_opt}

Using a similar probabilistic SLAM formulation as in Sec.~\ref{sec:Methodology_continuous_time_full_batch_opt}, we derive the cost function to minimize
\begin{align}\label{eq:cost_fun_full_batch_opti_dis_time}
    \min_{^{d}\mathcal{X}} & 
    \sum_{k = 1}^{K} \sum_{r \in \mathcal{R}_k} \norm{\mathbf{e}^{\text{v}}_{k,r}}^{2}_{\mathbf{W}_\text{v}} + 
    \sum_{k = 1}^{K} \norm{\mathbf{e}^{\text{i}}_{k}}^{2}_{\mathbf{W}_\text{i}} +
    \sum_{k = 1}^{K} (\norm{\mathbf{e}^{\text{b}_{\text{a}}}_{k}}^{2}_{\mathbf{W}_{\text{b}_{\text{a}}}} + \nonumber \\& \norm{\mathbf{e}^{\text{b}_{\omega}}_{k}}^{2}_{\mathbf{W}_{\text{b}_{\omega}}} ) +
    \sum_{d = 1}^{D} \norm{\mathbf{e}^{\text{gps}}_{d}}^{2}_{\mathbf{W}_\text{gps}},
\end{align}
The state vector is $^{d}\mathcal{X} = \{ ^{d}\mathcal{T}^{w}_{b}, \mathcal{V}^{w}_{b}, \mathcal{L}, t^{c}_{i}, \mathbf{T}^{c}_{i}, t^{g}_{i}, \mathbf{p}_{p}^{b}, ^{d}\mathcal{B}\}$.
The set $\mathcal{V}^{w}_{b}$ contains the velocity vectors: $\mathbf{v}^{w}_{b_k}$.
The set $^{d}\mathcal{B}$ contains the accelerometer and gyroscope bias vectors: $\mathbf{b}_{\text{a}_k}$ and $\mathbf{b}_{\omega_k}$.
The initial 3-D landmarks positions in $W$ are obtained 
similarly as described in~\ref{sec:Methodology_continuous_time_full_batch_opt}.
The reprojection errors $\mathbf{e}^{\text{v}}_{k,r}$ and the GPS errors $\mathbf{e}^{\text{gps}}_{d}$ are defined in the same way as in Sec.\ref{sec:Methodology_continuous_time_full_batch_opt} with the only difference that trajectory poses are represented in discrete time.
The errors $\mathbf{e}^{\text{i}}_{k}$ are the inertial residuals. 
We compute these residuals using the IMU preintegration formulation proposed in~\cite{Forster17troOnmanifold}.
Each inertial error $\mathbf{e}^{\text{i}}_{k}$ constrains the consecutive $k-1$ and $k$ poses and velocities according to the preintegrated IMU measurements in $[t_{k-1}, t_k]$ (see Eq. (45) in~\cite{Forster17troOnmanifold}).
The errors $\mathbf{e}^{\text{b}_{\text{a}}}_{k}$ and $\mathbf{e}^{\text{b}_{\omega}}_{k}$ constrain the bias random walk (see Eq. (48) in~\cite{Forster17troOnmanifold}).
\section{Experiments}\label{sec:Experiments}

We compare the continuous- and discrete-time representations in terms of accuracy of the estimated trajectory and time offsets.
To this end, we use the metrics~\cite{Zhang18iros}:
\begin{itemize}
    \item Positional absolute trajectory error ($\text{ATE}_{\text{P}}$) [m] : $\sqrt{\frac{1}{N} \sum_{k=0}^{N-1} \norm{ \mathbf{p}^w_{b_k} - \hat{\mathbf{p}}^w_{b_k}}^2 }$. 
    \item Rotational absolute trajectory error ($\text{ATE}_{\text{R}}$) [deg] : $\sqrt{\frac{1}{N} \sum_{k=0}^{N-1} \norm{ \text{Log}((\mathbf{R}^{w}_{b_k})^t \cdot \hat{\mathbf{R}}^w_{b_k} ) }^2 }$
\end{itemize}
In the case of continuous time, the trajectory is sampled at the timestamp of the camera measurements to obtain $\mathbf{p}^w_{b_k}$ and $\mathbf{R}^w_{b_k}$.
We run multiple experiments in hardware-in-the-loop simulation and on two real-world datasets.
The hardware-in-the-loop simulation allows a thorough evaluation of the capability of the two trajectory representations in estimating the time offsets since ground-truth values are known.
We investigate if the type of robot, flying or ground robot, has effects on the trajectory representation using the two real-world datasets.
We use the Ceres Solver~\cite{ceres-solver} and choose the Levenberg-Marquardt algorithm for the optimization.
Derivatives are computed using the automatic differentiation method included in the solver.
\begin{table}[H]
\caption{Ablation study on the order of the B-spline.
$\text{ATE}_{\text{P}}$ is in meters, $\text{ATE}_{\text{R}}$ is in degrees, time offset $t_i^c$ is in milliseconds.
In bold the best value for each sequence.}
\label{tab:euroc_ablation_study_spline_order}
\resizebox{0.49\textwidth}{!}{
\begin{tabular}{|c|c|c|c|c|c|c|c|}
\hline
\multirow{2}{*}{\textbf{\begin{tabular}[c]{@{}l@{}}Order\end{tabular}}} & \multirow{2}{*}{\textbf{\begin{tabular}[c]{@{}c@{}}Ev. \\ metric\end{tabular}}} & \multicolumn{6}{c|}{\textbf{EuRoC sequence}} \\ \cline{3-8} 
 & & \textbf{V101} & \textbf{V102} & \textbf{V103} & \textbf{V201} & \textbf{V202} & \textbf{V203} \\ \hline \hline 

\multirow{2}{*}{4} & $\text{ATE}_{\text{P}}$ [m] & \textbf{0.024} & \textbf{0.014} & \textbf{0.011} & \textbf{0.011} & 0.011 & 0.024 \\ & $\text{ATE}_{\text{R}}$ [deg] & \textbf{5.5} & \textbf{2.1} & \textbf{2.3} & \textbf{0.6} & 0.8 & 1.0 \\
 & $t_i^c$ [ms] & 0.9 & 3.2 & \textbf{-1.4} & 10.8 & 1.0 & 2.2 \\
\hline
\multirow{2}{*}{5} & $\text{ATE}_{\text{P}}$ [m] & \textbf{0.024} & \textbf{0.014} & \textbf{0.011} & \textbf{0.011} & \textbf{0.010} & 0.019 \\ & $\text{ATE}_{\text{R}}$ [deg] & \textbf{5.5} & 2.2 & \textbf{2.3} & 0.9 & \textbf{0.7} & 0.8 \\
 & $t_i^c$ [ms] & 0.3 & -5.8 & 2.0 & -1.8 & \textbf{0.0} & 0.5 \\
\hline
\multirow{2}{*}{6} & $\text{ATE}_{\text{P}}$ [m] & \textbf{0.024} & \textbf{0.014} & \textbf{0.011} & 0.012 & \textbf{0.010} & \textbf{0.010} \\ & $\text{ATE}_{\text{R}}$ [deg] & \textbf{5.5} & \textbf{2.1} & \textbf{2.3} & 0.8 & \textbf{0.7} & \textbf{0.6} \\
 & $t_i^c$ [ms] & \textbf{0.2} & \textbf{1.3} & \textbf{-1.4} & \textbf{1.2} & \textbf{0.0} & \textbf{0.2} \\
\hline
\end{tabular}
}
\end{table}
\begin{table}[H]
\caption{Ablation study on the frequency of the B-spline control nodes.
$\text{ATE}_{\text{P}}$ is in meters, $\text{ATE}_{\text{R}}$ is in degrees, estimated time offset $t_i^c$ is in milliseconds.
In bold the best value for each sequence.}
\label{tab:euroc_ablation_study_ct_nodes_freq}
\resizebox{0.49\textwidth}{!}{
\begin{tabular}{|c|c|c|c|c|c|c|c|}
\hline
\multirow{2}{*}{\textbf{\begin{tabular}[c]{@{}l@{}}Node \\ freq.\end{tabular}}} & \multirow{2}{*}{\textbf{\begin{tabular}[c]{@{}c@{}}Ev. \\ metric\end{tabular}}} & \multicolumn{6}{c|}{\textbf{EuRoC sequence}} \\ \cline{3-8} 
 & & \textbf{V101} & \textbf{V102} & \textbf{V103} & \textbf{V201} & \textbf{V202} & \textbf{V203} \\ \hline \hline 

\multirow{2}{*}{1 Hz} & $\text{ATE}_{\text{P}}$ [m] & 0.028 & 0.120 & 0.077 & 0.020 & 0.187 & 0.075 \\ & $\text{ATE}_{\text{R}}$ [deg] & 5.9 & 4.6 & 7.6 & 2.0 & 10.7 & 5.18 \\
 & $t_i^c$ [ms] & 6.6 & 19.3 & 1.2 & 0.0 & -45.6 & -19.0 \\
\hline
\multirow{2}{*}{5 Hz} & $\text{ATE}_{\text{P}}$ [m] & \textbf{0.023} & \textbf{0.014} & \textbf{0.011} & \textbf{0.010} & \textbf{0.010} & 0.019 \\ & $\text{ATE}_{\text{R}}$ [deg] & 5.6 & \textbf{2.1} & 2.3 & 0.9 & 0.8 & 0.8 \\
 & $t_i^c$ [ms] & -1.9 & -2.8 & 1.7 & 4.0 & 2.6 & -1.5 \\
\hline
\multirow{2}{*}{10 Hz} & $\text{ATE}_{\text{P}}$ [m] & 0.024 & \textbf{0.014} & \textbf{0.011} & 0.012 & \textbf{0.010} & \textbf{0.010} \\ & $\text{ATE}_{\text{R}}$ [deg] & \textbf{5.5} & \textbf{2.1} & 2.3 & \textbf{0.8} & \textbf{0.7} & \textbf{0.6} \\
 & $t_i^c$ [ms] & 0.2 & 1.3 & -1.4 & 1.2 & \textbf{0.0} & \textbf{0.2} \\
\hline
\multirow{2}{*}{20 Hz} & $\text{ATE}_{\text{P}}$ [m] & 0.025 & \textbf{0.014} & \textbf{0.011} & \textbf{0.010} & \textbf{0.010} & \textbf{0.010} \\ & $\text{ATE}_{\text{R}}$ [deg] & \textbf{5.5} & \textbf{2.1} & \textbf{2.2} & 1.2 & \textbf{0.7} & \textbf{0.6} \\
 & $t_i^c$ [ms] & -2.4 & \textbf{0.7} & \textbf{-0.9} & \textbf{-0.7} & -1.1 & 2.0 \\
\hline
\multirow{2}{*}{100 Hz} & $\text{ATE}_{\text{P}}$ [m] & 0.024 & 0.226 & 0.117 & 0.060 & 0.168 & 0.136 \\ & $\text{ATE}_{\text{R}}$ [deg] & 8.8 & 8.4 & 12.1 & 6.6 & 11.1 & 5.6 \\
 & $t_i^c$ [ms] & \textbf{0.0} & -4.1 & -3.0 & 0.0 & -1.3 & -0.5 \\
\hline
\end{tabular}
}
\end{table}
\noindent
We ran all the experiments on Ubuntu 18.04 workstation with an Intel Core i7 3.2GHz Processor and used 8 cores for the optimization.
In all the experiments, the optimization problem is solved until convergence. 
This is always achieved in less than 50 iterations.

\subsection{Hardware-in-the-Loop Simulation: EuRoC Dataset}\label{sec:Experiments_EuRoC_Dataset}

The EuRoC dataset~\cite{Burri15ijrr} contains sequences recorded on-board a hex-rotor flying robot equipped with a stereo camera and an IMU.
This dataset is well-known for accurate ground-truth and hardware synchronizated sensors. 
The camera-IMU time offset can be assumed to be zero. 
We only use the sequences labeled with V\_, which contain 6-DOF ground-truth from a motion capture system. 
Sequences are classified (see Table 2 in~\cite{Burri15ijrr}) as easy, medium, or hard, reflecting the level of difficulty due to illumination conditions, scene texture and vehicle motion.
Difficult sequences contain challenging illumination conditions, (e.g., motion blur), and fast motions with average linear and angular velocities up to 0.9 m$\text{s}^{-1}$ and 0.75 rad $\text{s}^{-1}$, respectively.
\begin{table*}[t]
\caption{Comparison of continuous- and discrete-time approaches on the EuRoC dataset. 
$\text{ATE}_{\text{P}}$ is in meters, $\text{ATE}_{\text{R}}$ is in degrees, $\hat{t}_i^c$ (ground-truth) and $t_i^c$ (estimated) time offset are in milliseconds.
The best values of $\text{ATE}_{\text{P}}$ and $\text{ATE}_{\text{R}}$ for each sequence are in bold.}
\centering
\label{tab:euroc_comparison_traj_representation}
\resizebox{1.0\textwidth}{!}{
\begin{tabular}{|c|cc|cc|cc|cc|cc|cc|}
\hline
\multirow{2}{*}{\textbf{Seq.}} & \multicolumn{6}{c|}{\textbf{Continuous-time}} & \multicolumn{6}{c|}{\textbf{Discrete-time}} \\ \cline{2-13} 
 & \multicolumn{2}{c|}{$\hat{t}^c_i$ = 0 {[}ms{]}} & \multicolumn{2}{c|}{$\hat{t}^c_i$ = 10 {[}ms{]}} & \multicolumn{2}{c|}{$\hat{t}^c_i$ = 20 {[}ms{]}} & \multicolumn{2}{c|}{$\hat{t}^c_i$ = 0 {[}ms{]}} & \multicolumn{2}{c|}{$\hat{t}^c_i$ = 10 {[}ms{]}} & \multicolumn{2}{c|}{$\hat{t}^c_i$ = 20 {[}ms{]}} \\ \cline{2-13} 
 & $\text{ATE}_{\text{P}}$ / $\text{ATE}_{\text{R}}$  & $t_i^c$  & $\text{ATE}_{\text{P}}$  / $\text{ATE}_{\text{R}}$  & $t_i^c$  & $\text{ATE}_{\text{P}}$  / $\text{ATE}_{\text{R}}$  & $t_i^c$  & $\text{ATE}_{\text{P}}$  / $\text{ATE}_{\text{R}}$  & \multicolumn{1}{c|}{$t_i^c$ } & $\text{ATE}_{\text{P}}$  / $\text{ATE}_{\text{R}}$  & \multicolumn{1}{c|}{$t_i^c$ } & $\text{ATE}_{\text{P}}$  / $\text{ATE}_{\text{R}}$  & $t_i^c$  \\
\hline \hline
V101 & 0.024 / \textbf{5.5} & 0.2 & 0.024 / \textbf{5.5} & 11.0 & 0.024 / \textbf{5.5} & 22.2 & \textbf{0.016} / 5.6 & \multicolumn{1}{c|}{0.3} & \textbf{0.016} / 5.6 & \multicolumn{1}{c|}{9.1} & \textbf{0.016} / 5.6 &  18.6\\
V102 & \textbf{0.014} / \textbf{2.1} & 1.3 & \textbf{0.014} / \textbf{2.1} & 9.7 & \textbf{0.014} / \textbf{2.1} & 20.6 & 0.024 / 2.4 &  \multicolumn{1}{c|}{0.0} & 0.026 / 2.3 & \multicolumn{1}{c|}{4.6} & 0.031 / 2.2 &  9.3\\
V103 & \textbf{0.011} / \textbf{2.3} & -1.4 & \textbf{0.011} / \textbf{2.3} & 11.8 & \textbf{0.011} / \textbf{2.3} & 22.3 & 0.018 / 2.7 &  \multicolumn{1}{c|}{0.0} & 0.020 / 2.6 & \multicolumn{1}{c|}{3.5} & 0.024 / 2.6 &  7.2\\
V201 & 0.012 / \textbf{0.8} & 1.2 & 0.010 / 0.9 & 9.7 & 0.010 / 0.9 & 19.0 & \textbf{0.009} / 1.0 &  \multicolumn{1}{c|}{0.3} & 0.010 / 1.0 & \multicolumn{1}{c|}{8.1} & 0.012 / 1.0 &  16.4\\
V202 & \textbf{0.010} / \textbf{0.7} & 0.0 & \textbf{0.010} / \textbf{0.7} & 10.0 & \textbf{0.010} / \textbf{0.7} & 20.0 & 0.019 / 0.8 &  \multicolumn{1}{c|}{0.0} & 0.021 / 0.9 & \multicolumn{1}{c|}{8.5} & 0.024 / 1.1 &  16.7\\
V203 & \textbf{0.010} / \textbf{0.6} & 0.2 & \textbf{0.010} / \textbf{0.6} & 10.6 & \textbf{0.010} / \textbf{0.6} & 21.5 & 0.033 / 1.1 &  \multicolumn{1}{c|}{0.0} & 0.036 / 1.2 & \multicolumn{1}{c|}{4.0} & 0.040 / 1.3 & 7.5 \\
\hline
\end{tabular}
}
\end{table*}
\begin{table}[t]
\caption{Full-batch optimization time. For the continuous-time representation, we used B-splines of order 6 and control frequency $\in [1, 2, 10, 20, 100]$ [Hz]. The time values are in minutes. The trajectory length is in meters. The lowest time per sequence is in bold.}
\label{tab:euroc_timing}
\centering
\resizebox{0.49\textwidth}{!}{
\begin{tabular}{|c|c|cllll|c|}
\hline
{\multirow{2}{*}{\textbf{Seq.}}} &
  {\multirow{2}{*}{\textbf{\begin{tabular}[c]{@{}c@{}}Length \\ \text{[m]}\end{tabular}}}} &
  \multicolumn{5}{c|}{\textbf{CT}} &
  \multirow{2}{*}{\textbf{DT}} \\ \cline{3-7}
     &      & \multicolumn{1}{c|}{\textbf{1 Hz}} & \multicolumn{1}{c|}{\textbf{2 Hz}} & \multicolumn{1}{c|}{\textbf{10 Hz}} & \multicolumn{1}{c|}{\textbf{20 Hz}} & \textbf{100 Hz} &  \\ \hline \hline
V101 & 52.6 & \multicolumn{1}{c|}{13}  & \multicolumn{1}{c|}{15}  & \multicolumn{1}{c|}{21}   & \multicolumn{1}{c|}{36}   & \multicolumn{1}{c|}{37}    &  \textbf{12} \\
V102 & 75.9 & \multicolumn{1}{c|}{\textbf{19}}  & \multicolumn{1}{c|}{20}  & \multicolumn{1}{c|}{25}   & \multicolumn{1}{c|}{49}   & \multicolumn{1}{c|}{57} & 28 \\
V103 & 79.0 & \multicolumn{1}{c|}{24}  & \multicolumn{1}{c|}{37}  & \multicolumn{1}{c|}{22}   & \multicolumn{1}{c|}{46}   & \multicolumn{1}{c|}{46} & \textbf{21} \\
V201 & 36.5 & \multicolumn{1}{c|}{6}  & \multicolumn{1}{c|}{\textbf{5}}  & \multicolumn{1}{c|}{13} & \multicolumn{1}{c|}{10}   &  \multicolumn{1}{c|}{13} & 15 \\
V202 & 83.2 & \multicolumn{1}{c|}{44}  & \multicolumn{1}{c|}{54}  & \multicolumn{1}{c|}{73} & \multicolumn{1}{c|}{44} & \multicolumn{1}{c|}{77}  & \textbf{21} \\
V203 & 86.1 & \multicolumn{1}{c|}{18} & \multicolumn{1}{c|}{16} & \multicolumn{1}{c|}{39} & \multicolumn{1}{c|}{50} & \multicolumn{1}{c|}{31} & \textbf{14} \\ \hline
\end{tabular}
}
\end{table}
\noindent
We simulate GPS measurements by downsampling and corrupting the ground-truth positions with zero-mean Gaussian noise.
The rate of the simulated GPS measurements is 10 Hz and the standard deviation of the Gaussian noise is $0.1$ m.
The position offset $\mathbf{p}^b_p$ is equal to $\mathbf{0} \in \mathbb{R}^3$.
We only use images from the right camera.

\subsubsection{Ablation study on the B-spline}\label{sec:euroc_ablation_study_on_bspline}

We conducted a study to evaluate the effects of the order and frequency of the control points of the B-spline on the trajectory and camera-IMU time offset estimates.
The initial value of the time offset was set to 0 ms.
The results of the ablation study on the order of the B-spline are in Table~\ref{tab:euroc_ablation_study_spline_order}.
A B-spline of order 6, which results in a cubic polynomial encoding accelerations, is needed to correctly estimate the camera-IMU time offset.
This conclusion agrees with the findings in~\cite{Furgale13iros}.
An order higher than 6 does not have any effect on the estimation results.
When the spline order is 6, the values of the time offset are close but not exactly 0 ms.
Since we solve the optimization till convergence, the solution has reached a local minimum of the cost function.
How far is the local minimum from the global minimum is unknown,
and, in general, even the unknown global optimum of the MAP estimation can be different from the ground-truth due to modeling errors.

We set the order of the spline to 6 and performed the ablation study on the control node frequency. The results are in Table~\ref{tab:euroc_ablation_study_ct_nodes_freq}.
The values of ATE and $t^{c}_{i}$ suggest that a frequency of 10 Hz is the optimal choice. Increasing to 20 Hz does not give any significant advantages, while making the optimization computationally more expensive.
When the frequency is high, e.g., 100 Hz, the convergence to a good solution is hard to achieve.
We conclude that order 6 and control nodes frequency 10/20 Hz are appropriate parameters for B-spline representing trajectories in this dataset.

\subsubsection{Evaluation of the trajectory representation}

In this set of experiments, we evaluated the continuous- and discrete-time trajectory representations in terms of $\text{ATE}_{\text{P}}$, $\text{ATE}_{\text{R}}$, accuracy in estimating the camera-IMU time offset, and computational cost.
Following the findings of Sec.~\ref{sec:euroc_ablation_study_on_bspline}, we used B-spline of order 6 and control node frequency of 10 Hz.
To evaluate the accuracy in estimating the camera-IMU time offset, we simulated delays in the camera data stream, similarly to~\cite{qin2018online}.
We ran experiments with delays of 0, 10, and 20 ms. The results of this comparison are in Table~\ref{tab:euroc_comparison_traj_representation}.
These results suggest that when the camera and IMU are time-synchronized both trajectory representations produce similar accuracy (see the column corresponding to $t^c_i = 0$).
However, using continuous time gives a lower $\text{ATE}_{\text{P}}$ in the sequences V102, V103, V202, and V203, which are medium and difficult sequences containing fast motion of the robot.

When the camera and IMU are not time-synchronized, continuous time is the best trajectory representation.
This representation can properly estimate the time offset and produces $\text{ATE}$ similar to the case of time-synchronized sensors.
In particular, the discrete-time representation suffers in estimating the camera-IMU time offset in fast trajectories.
This can be seen from the large difference between the estimated and the true time offset in the sequences V102, V103, V202, and V203.
The reason for this result is the assumption of camera motion with constant velocity in the period of time between consecutive camera frames, which is needed to compute the velocity of the 2-D features as described in Sec.~\ref{sec:Methodology_Discrete_time}.
For agile motion, this assumption is no longer accurate.
Table~\ref{tab:euroc_timing} contains the optimization time, which is the time that the solver takes to converge.

\subsubsection{Ablation study on the contribution of the different sensor modalities}

In this section, we are interested in studying the contributions of each sensor modality, i.e., vision, inertia and GPS, to the accuracy of the trajectory estimation.
To this end, we solved the optimization problems in Eq.~(\ref{eq:cost_fun_full_batch_opti_cont_time}) and Eq.~(\ref{eq:cost_fun_full_batch_opti_dis_time}) for the continuous- and discrete-time case respectively, with all possible combinations of at least two sensors.
For the continuous-time representation, we used B-splines of order 6 with control node frequency of 10 Hz.
The values of $\text{ATE}_{\text{P}}$ and $\text{ATE}_{\text{R}}$ for this analysis are in Table~\ref{tab:euroc_ablation_study_sensors_contribution_ate_p} and Table~\ref{tab:euroc_ablation_study_sensors_contribution_ate_r}.
The results show that vision is the most important sensor modality.
For both trajectory representations, the vision-GPS configuration produces the lowest $\text{ATE}_{\text{P}}$ in most of the sequences.
In this case, the COLMAP reconstructed trajectory is transformed to the gravity aligned frame by using noisy simulated GPS measurements and then re-optimized together with those measurements.
COLMAP is able to reconstruct an accurate trajectory on every sequence of this dataset.
For this reason, including the inertial measurements in the estimation process does not bring any significant benefit.
\begin{table}[H]
\caption{Ablation study on the contribution of the different sensor modalities. The error metric is the $\text{ATE}_{\text{P}}$ in meters.
V: vision, I: inertia, G: simulated GPS.
In bold the best value for each sequence.}
\label{tab:euroc_ablation_study_sensors_contribution_ate_p}
\begin{tabular}{|c|c|c|c|c|c|c|c|}
\cline{1-8}
\multirow{2}{*}{\textbf{Sensors}} & \multirow{2}{*}{\textbf{\begin{tabular}[c]{@{}l@{}}Traj. \\ repr.\end{tabular}}} & \multicolumn{6}{c|}{\textbf{EuRoC Sequence}} \\ \cline{3-8} 
{} & {} & \textbf{V101} & \textbf{V102} & \textbf{V103} & \textbf{V201} & \textbf{V202} & \textbf{V203} \\ 
\hline \hline
{\multirow{2}{*}{V+I+G}} & CT & 0.024 & 0.014 & \textbf{0.011} & 0.012 & 0.010 & \textbf{0.010} \\
{} & DT & 0.016 & 0.024 & 0.018 & \textbf{0.009} & 0.018 & 0.033 \\ \cline{1-2} \hline
{\multirow{2}{*}{V+G}} & CT &  0.011 & \textbf{0.013} & 0.012 & \textbf{0.009} & \textbf{0.008} & 0.012 \\
{} & DT & \textbf{0.010} & 0.025 & 0.024 & 0.010 & 0.012 & 0.029 \\ \cline{1-2} \hline
{\multirow{2}{*}{I+G}} & CT & 0.062 & 0.102 & 0.117 & 0.112 & 0.164 & 0.363 \\
{} & DT & 0.139 & 0.137 & 0.138 & 0.138 & 0.138 & 0.139 \\ \cline{1-2} \hline
{\multirow{2}{*}{V+I}} & CT & 0.039 & 0.022 & 0.014 & 0.065 & 0.030 & 0.015 \\
{} & DT & 0.081 & 0.106 & 0.030 & 0.064 & 0.022 & 0.031 \\
\hline
\end{tabular}
\end{table}
\begin{table}[H]
\caption{Ablation study on the contribution of the different sensor modalities. The error metric is $\text{ATE}_{\text{R}}$ in degrees.
V: vision, I: inertia, G: simulated GPS.
In bold the best value for each sequence.}
\label{tab:euroc_ablation_study_sensors_contribution_ate_r}
\begin{tabular}{|c|c|c|c|c|c|c|c|}
\cline{1-8}
{\multirow{2}{*}{\textbf{Sensors}}} & {\multirow{2}{*}{\textbf{\begin{tabular}[c]{@{}l@{}}Traj. \\ repr.\end{tabular}}}} & \multicolumn{6}{c|}{\textbf{EuRoC Sequence}} \\ \cline{3-8} 
{} & {} & \textbf{V101} & \textbf{V102} & \textbf{V103} & \textbf{V201} & \textbf{V202} & \textbf{V203} \\ 
\hline \hline
{\multirow{2}{*}{V+I+G}} & CT & 5.5 & 2.1 & 2.3 & \textbf{0.8} & \textbf{0.7} & \textbf{0.6} \\
{} & DT & 5.6 & 2.4 & 2.7 & 1.0 & 0.8 & 1.1 \\ \cline{1-2} \hline
{\multirow{2}{*}{V+G}} & CT &  5.5 & 2.3 & 2.5 & 1.1 & 0.8 & 1.2 \\
{} & DT & 5.6 & 2.3 & 2.7 & 1.1 & 0.8 & 0.9 \\ \cline{1-2} \hline
{\multirow{2}{*}{I+G}} & CT & 10.5 & 6.3 & 7.6 & 11.0 & 11.0 & 9.4 \\
{} & DT & 12.3 & 7.7 & 8.8 & 11.8 & 11.8 & 10.6 \\ \cline{1-2} \hline
{\multirow{2}{*}{V+I}} & CT & \textbf{5.4} & \textbf{2.0} & \textbf{2.2} & 0.9 & 0.8 & \textbf{0.6} \\
{} & DT & 5.5 & 2.2 & 2.8 & 1.0 & 0.8 & 1.2 \\
\hline
\end{tabular}
\end{table}

\subsection{Actual GPS with an outdoor flying robot}\label{sec:Experiments_Outdoor_Dataset_Flying_Robot}

\begin{figure}[t]
    \centering
    \includegraphics[width=1.0\linewidth]{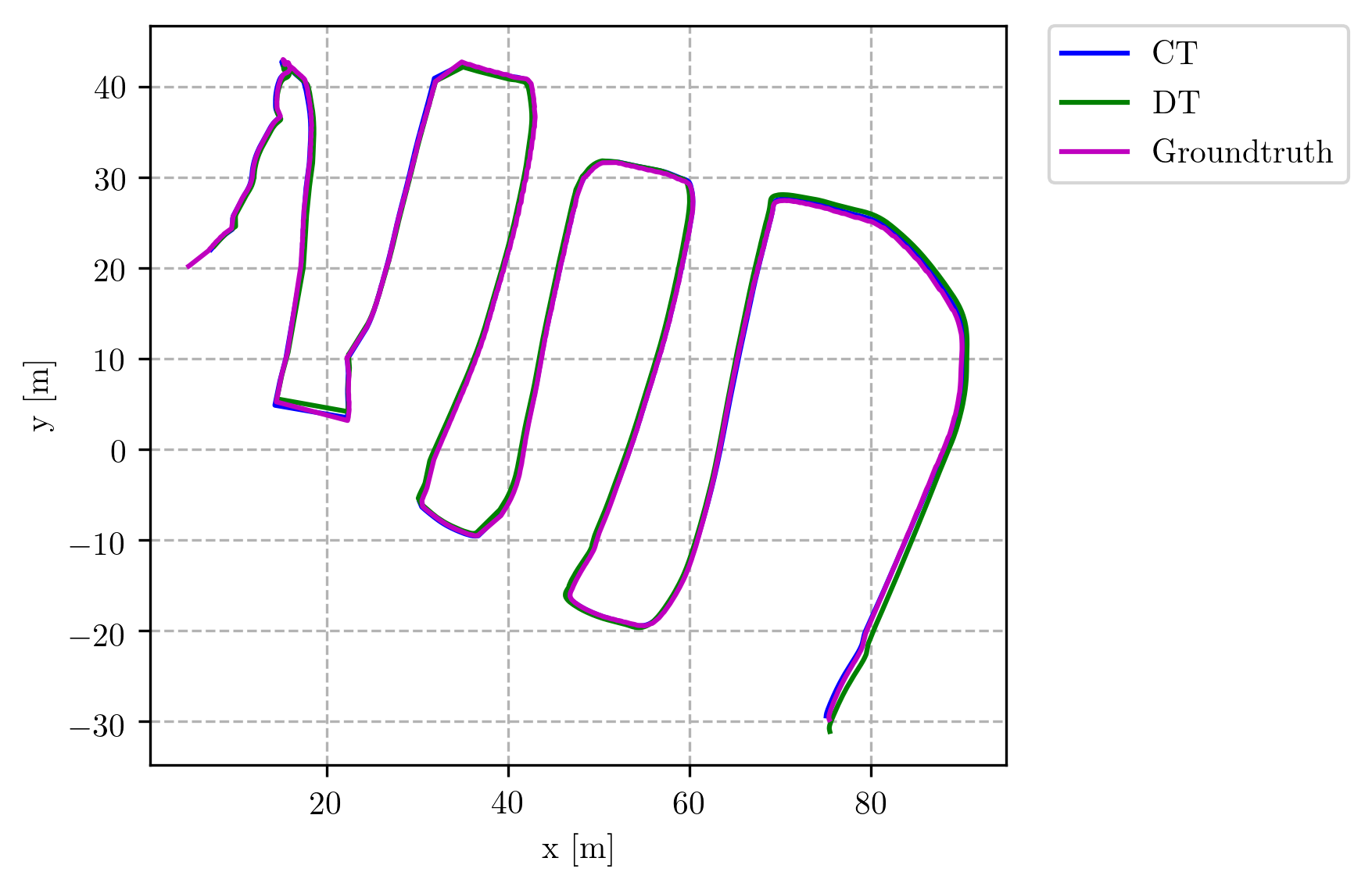}
    \caption{$XY$-view of the trajectory flown in Sequence 1 of the outdoor flying robot experiments.}
    \label{fig:flyingrobot_xytraj}
\end{figure}

This dataset, courtesy of~\cite{Surber17icra}, contains outdoor flights of a flying robot equipped with a time-synchronized VI sensor (stereo camera and IMU), and a GPS receiver.
GPS measurements are available at 5 Hz.
We only use images from the left camera.
The ground-truth position is provided by a Leica total station.
Fig.~\ref{fig:flyingrobot_xytraj} shows the first trajectory of the dataset.
The $\text{ATE}_{\text{P}}$ and the time offset estimates are in Table~\ref{tab:flying_robot_dataset}.
For the continuous-time case, we used B-splines of order 6 and control node frequency of 10 Hz. 
As suggested by the results in Sec.~\ref{sec:Experiments_EuRoC_Dataset}, these values are suited for encoding flying robot trajectories.
These results confirm the findings of Sec.~\ref{sec:Experiments_EuRoC_Dataset}: when the sensor are time-synchronized the two trajectory representations produce similar results, as shown by the similar values of $\text{ATE}_{\text{P}}$ and camera-IMU time offset.

\subsection{Outdoor Trajectory: Ground robot}\label{sec:Experiments_Outdoor_Dataset_Ground_Robot}
\begin{figure}[t]
    \centering
    \includegraphics[width=1.0\linewidth]{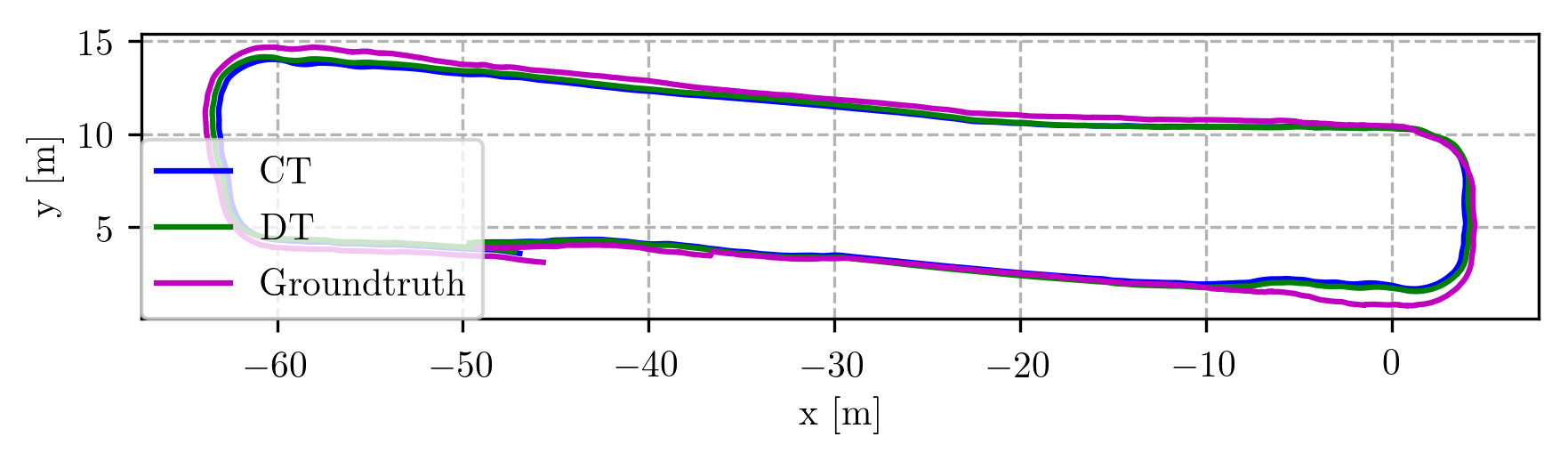}
    \caption{$XY$-view of the trajectory traveled by the ground robot.}
    \label{fig:groundedgrobot_xytraj}
\end{figure}

This experiment contains an evaluation of a trajectory recorded on-board a ground robot.
The robot is equipped with a time-synchronized VI sensor (monocular camera and IMU) and a GPS antenna~\footnote{\url{https://www.fixposition.com/}}.
The GPS measurements are available at 5 Hz.
The 3-D position ground-truth is provided by a RTK-GPS system.
Fig.~\ref{fig:groundedgrobot_xytraj} shows the traveled trajectory of the robot.
To study the influence of the B-spline order and control point frequency, we ran experiments with orders: 5, 6, and 7, and control node frequency: 10, 20, 100 Hz.
Both continuous-time and discrete-time representations produce similar $\text{ATE}_{\text{P}}$ as reported in Table~\ref{tab:ground_robot_dataset}.
The estimated time offsets are similar for all the configurations listed in Table~\ref{tab:ground_robot_dataset}.
\begin{table}[t]
\caption{Comparison of continuous- and discrete-time approaches on the outdoor flying robot dataset.
$\text{ATE}_{\text{P}}$ is in meters and $t^c_i$ is in milliseconds.
The best values of $\text{ATE}_{\text{P}}$ for each sequence are in bold.}
\label{tab:flying_robot_dataset}
\centering
\resizebox{0.35\textwidth}{!}{
\begin{tabular}{|c|c|c|c|}
\cline{1-4}
\textbf{Err. metric} & \textbf{Traj. repr.} & \textbf{Seq. 1} & \textbf{Seq. 2} \\
\hline \hline
\multirow{2}{*}{$\textbf{ATE}_{\text{P}}$ \textbf{[m]}} & \textbf{CT} & \textbf{0.39} &  \textbf{0.50}\\
{} & \textbf{DT} & {0.60} & 0.86 \\ \cline{1-4}
\multirow{2}{*}{$\mathbf{t^c_i}$ \textbf{[ms]}} & \textbf{CT} & {0.4} & 0.6 \\
{} & \textbf{DT} & {0.4} & 0.4 \\ \cline{1-4}
\multirow{2}{*}{$\mathbf{t^g_i}$ \textbf{[ms]}} & \textbf{CT} & {-87.0} &  -118.0\\
{} & \textbf{DT} & {-81.0} & -119.0 \\
\hline
\end{tabular}
}
\end{table}
\noindent
In the continuous-time case, with B-spline of order 6 and control node frequency of 10 Hz, $t^c_i$, and $t^{g}_i$ are -1.5 ms, and -26.0 ms, respectively.
For the discrete-time case, they are -0.8 ms, and -36.3 ms.
These results show that the findings of the experiments with a flying robot also apply to the case of a ground robot.
\begin{table}[t]
\caption{Comparison of continuous- and discrete-time approaches on the outdoor ground robot dataset. 
The error metric is the $\text{ATE}_{\text{P}}$ in meters.
The best value is in bold.}
\label{tab:ground_robot_dataset}
\centering
\resizebox{0.28\textwidth}{!}{
\begin{tabular}{|c||cccc|}
\hline
\multirow{5}{*}{\textbf{CT}} &
  \multicolumn{1}{c|}{\multirow{2}{*}{\textbf{\begin{tabular}[c]{@{}c@{}}Freq.\\ {[}Hz{]}\end{tabular}}}} &
  \multicolumn{3}{c|}{\textbf{Order}} \\ \cline{3-5} 
            & \multicolumn{1}{c|}{}       & \multicolumn{1}{c|}{\textbf{5}} & \multicolumn{1}{c|}{\textbf{6}} & \textbf{7}    \\ \cline{2-5}
            & \multicolumn{1}{c|}{10}  & \multicolumn{1}{c|}{0.93} & \multicolumn{1}{c|}{0.92} & 0.93 \\ \cline{2-5} 
            & \multicolumn{1}{c|}{20}  & \multicolumn{1}{c|}{1.01} & \multicolumn{1}{c|}{0.95} & 0.99 \\ \cline{2-5} 
            & \multicolumn{1}{c|}{100} & \multicolumn{1}{c|}{0.80} & \multicolumn{1}{c|}{0.89} & \textbf{0.78} \\ \hline \hline
\textbf{DT} & \multicolumn{4}{c|}{0.87}                                                                  \\ \hline
\end{tabular}
}
\end{table}
\section{Conclusions} \label{sec:Conclusions}

The objective of this work is to compare continuous vs. discrete vision-based SLAM formulations to guide practitioners in the development of SLAM algorithms.
We performed ablation and comparative studies in a hardware-in-the-loop simulation with full knowledge of the ground-truth.
The ablation studies on the order and frequency of the control nodes of the B-splines suggest that it is necessary to use B-splines of order 6 and control node frequency of at least 10 Hz to accurately estimate the camera-IMU time offset.
The comparative studies aimed at comparing continuous- and discrete-time trajectory representations with different levels of camera-IMU time delay.
We find that when the camera and IMU are time-synchronized the two representations produce similar results.
When a delay is present between the two measurement streams, the continuous-time representation is able to recover an accurate estimate of the time offset and consequently, produces lower ATE.
In contrast, the discrete-time formulation fails in estimating the time offset, particularly when the robot moves fast, which consequently leads to high values of the ATE.
The main reason of this result is that the assumption, which is necessary to estimate the camera-IMU time offset, of constant velocity of the camera motion in the period of time between consecutive camera frames does not always hold.
The findings of the hardware-in-the-loop simulation agree with the results of the experiments on the real-world datasets containing data from aerial and ground robots.
In addition, we evaluated the contribution of each sensor modality in an ablation study.
We found that on the EuRoC dataset the most important sensor modality is vision.
In most of the sequences, the lowest $\text{ATE}_{\text{P}}$ is obtained by aligning the camera trajectory obtained from COLMAP to a gravity aligned frame by using the noisy simulated GPS measurements.
Including inertial measurements does not give any significant advantage.
\section{Acknowledgments} \label{sec:Acknowledgments}

The authors would like to thank Torsten Sattler and Antonio Loquercio for the fruitful discussions, and the team at Fixposition for the ground robot dataset.

{\small
\bibliographystyle{IEEEtran}
\bibliography{all}
}

\end{document}